\documentclass[sigconf]{acmart}

\usepackage{booktabs} 
\usepackage{algpseudocode}
\usepackage{algorithm} 
\definecolor{light-gray}{gray}{0.92}

\algrenewcommand{\algorithmiccomment}[1]{\hfill\colorbox{light-gray}{$\triangleright$ #1}}

\makeatletter
\newcommand{\Colorbox}[1]{\colorbox{light-gray}{#1}}

\makeatother

\setcopyright{rightsretained}

\let\oldbibitem\bibitem
\def\bibitem{\vfill\oldbibitem}
\usepackage{setspace}
\AtBeginDocument{%
  \addtolength\abovedisplayskip{-0.25\baselineskip}%
  \addtolength\belowdisplayskip{-0.25\baselineskip}%
}

\usepackage{bm}
\usepackage{dsfont}

\setcopyright{rightsretained}
\copyrightyear{2020}
\acmYear{2020}
\setcopyright{rightsretained}
\acmConference[GECCO '20]{Genetic and Evolutionary Computation Conference}{July 8--12, 2020}{Canc\'n, Mexico}
\acmBooktitle{Genetic and Evolutionary Computation Conference (GECCO '20), July 8--12, 2020, Canc\'un, Mexico}\acmDOI{10.1145/3377930.3390203}
\acmISBN{978-1-4503-7128-5/20/07}

\begin{document}
\title{Quality Diversity for Multi-task Optimization}

\author{Jean-Baptiste Mouret}
\orcid{0000-0002-2513-027X}
\affiliation{%
  \institution{Inria, CNRS, Universit\'e de Lorraine}
  \streetaddress{615 rue du Jardin Botanique}
  \city{Nancy}
  \state{France}
  \postcode{54600}
}
\email{jean-baptiste.mouret@inria.fr}

\author{Glenn Maguire}
\affiliation{%
  \institution{Inria, CNRS, Universit\'e de Lorraine}
  \streetaddress{615 rue du Jardin Botanique}
  \city{Nancy}
  \state{France}
  \postcode{54600}
}
\email{glenn.maguire@inria.fr}


\begin{abstract}
Quality Diversity (QD) algorithms are a recent family of optimization algorithms that search for a large set of diverse but high-performing solutions. In some specific situations, they can solve multiple tasks at once. For instance, they can find the joint positions required for a robotic arm to reach a set of points, which can also be solved by running a classic optimizer for each target point. However, they cannot solve multiple tasks when the fitness needs to be evaluated independently for each task (e.g., optimizing policies to grasp many different objects). In this paper, we propose an extension of the MAP-Elites algorithm, called Multi-task MAP-Elites, that solves multiple tasks when the fitness function depends on the task. We evaluate it on a simulated parameterized planar arm (10-dimensional search space; 5000 tasks) and on a simulated 6-legged robot with legs of different lengths (36-dimensional search space; 2000 tasks). The results show that in both cases our algorithm outperforms the optimization of each task separately with the CMA-ES algorithm.

\end{abstract}

%
%


\maketitle

\section{Introduction}
Quality Diversity (QD) algorithms are a recent family of optimization algorithms that search for a large set of diverse but high-performing solutions \cite{mouret2015illuminating,cully2017quality,pugh2016quality}, instead of the global optimum, like in single-objective optimization, or the Pareto frontier, like in multi-objective optimization. For instance, when optimizing aerodynamic 3D shapes, a user might want to be presented with multiple low-drag solutions of diverse materials and curvatures, and then select the best one according to criteria that are not encoded in the fitness function, such as aesthetics \cite{gaier2018}

So far, QD algorithms have very promising results for at least robotics, engineering, and video games. For example, a QD algorithm was used to evolve repertoires of diverse gaits for legged robots~\cite{cully:hal-01158243,chatzilygeroudis:hal-01654641,duarte2017evolution,kaushik2020}; in computer-aided design, similar algorithms were used to propose shapes of high-speed bicycles with various curvatures and volumes \cite{gaier2018}; in artificial intelligence for games research, QD algorithms have been implemented to generate procedural content like game levels or spaceships \cite{gravina2019procedural}.

In essence, QD algorithms solve multiple tasks at once. For instance, a QD algorithm that evolves gaits to reach each point in the plane is actually solving the task ``how to reach ($x$, $y$)?'', for each value of $x$ and $y$ within their respective bounds: each of these tasks could be solved separately with a single-objective optimization and a fitness function that computes the distance between the achieved position and the target. The ``magic'' of QD algorithms is that they reuse solutions that might have been unfit for some tasks (e.g., we want to reach ($x_1$, $y_1$) but we arrived at ($x_2, y_2$)) to solve another task (e.g., reaching ($x_2, y_2$)).

However, these algorithms require the family of tasks to have a very specific structure, namely, that both the features of the candidate solutions (i.e., the specification of the task) and the fitness values can be obtained from a single call to the fitness function. While this is the case for some tasks, like walking in different ways (we can measure the features of the gaits and the fitness during the simulation) or evolving shapes of different volumes (we can measure both the fitness and the volume from a generated candidate), there are many families of tasks for which a different fitness call is required to know the performance of a candidate for each task. For example, we might want to find a diverse set of policies or trajectories to address:
\begin{itemize}
    \item Walking gaits that work for a family of damage conditions --- the best gait with a missing front leg; the best gait with a shortened back leg; etc. In this case, we need to simulate the robot under each of these conditions separately in order to determine the performance for each task\footnote{Please note that this is different from finding many gaits for the same morphology, like in \cite{cully:hal-01158243}, which can be achieved with a traditional QD algorithm: such gaits are often useful for different morphology (e.g., for damage recovery), but they are not explicitly selected for this.}.
    \item Grasping many different objects, each of them with a different grasping policy. Here we also need to simulate the grasping of a each object to know how well particular a set of parameters will perform.
    \item Successfully completing all the levels of a video game, which requires playing each level in order to determine how well a given policy performs on a given level.
\end{itemize}
Current QD algorithms cannot solve this kind of multi-task challenge without evaluating the fitness function on all the tasks, which requires a prohibitively large number of evaluations when the number of tasks is large (e.g., more than a few dozen).

In this paper, we propose an extension of the MAP-Elites algorithm \cite{mouret2015illuminating} that solves numerous tasks simultaneously (more than a few thousands) when the fitness function must evaluated separately for each task. The key intuition in our approach is that a high-performing solution for a task is likely to be a good starting point to solve another task of the same family. This means that solving all the tasks together, like in QD algorithms, should be faster than solving each of them independently. This similarity between solutions for different tasks (or niches) was studied in previous work in QD \cite{vassiliades:hal-01764739}, which led to the conclusion that high-performing solutions occupy a specific hypervolume of the genotype space in spite of being evenly spread in the diversity space. Stated differently, the high-performing solutions that QD algorithms find are different --- by design --- but are similar in the genotypic space. We hypothetize that a similar effect happens when solving many tasks simultaneously.

Compared to the MAP-Elites algorithm \cite{mouret2015illuminating}, this paper proposes an algorithm to select the task, that is, the right ``niche'' (or cell) on which to evaluate each of the candidate solutions that are generated by the variation operator. This task selection operator uses the distance between the tasks (e.g., the distance between the parameters of two tasks) to exploit their similarities. The rest of the algorithm is the same as the vanilla MAP-Elites except that behavioral descriptors are replaced by task descriptors, which are chosen beforehand instead of being extracted from the evaluations.


\section{Related Work}
\subsection{Quality diversity algorithms}
Quality diversity algorithms are descendants of Novelty Search~\cite{lehman2011abandoning}, behavioral diversity maintaintance~\cite{mouret:hal-00473132,mouret:hal-00687609,mouret:hal-00473147} and niching methods like fitness sharing \cite{sareni1998fitness}. Like Novelty Search and behavioral diversity, they define diversity in the ``behavioral space'' (or ``feature space''), which is defined by descriptors of the features of each candidate solution observed during the evaluation of the fitness \cite{lehman2011abandoning, mouret:hal-00473132,mouret:hal-00687609,mouret:hal-00473147}. For instance, two robots might cover the same distance but with a different trajectory: if the objective is to go as far as possible, these two robots would have the same fitness but different behavioral descriptors.

QD algorithms fall into two categories \cite{cully2017quality}: population-based algorithms and archive-based algorithms. Both kinds use an archive to store previously encountered solutions. In population-based algorithms, the archive is used to steer a population towards new parts of the search space. For example, in Novelty Search with Local Competition \cite{lehman2011evolving}, a multi-objective optimization algorithm (NSGA-II) ranks individuals according to two objectives: their fitness relative to their behavioral neighbor in the archive and the mean distance to individuals already in the archive. By contrast, in archive-based algorithms, the archive \emph{is} the population: new individuals are created by selecting parents from the archive and applying the variation operators (mutation and cross-over).

MAP-Elites \cite{mouret2015illuminating} is one of the most used archive-based algorithms because it is conceptually simple and leads to good results \cite{cully:hal-01158243,chatzilygeroudis:hal-01654641,duarte2017evolution,kaushik2020,gravina2019procedural,gaier:hal-01518698}. MAP-Elites divides the feature space into niches (or bins) according to a regular grid \cite{mouret2015illuminating} or a centroidal Voronoi tesselation \cite{vassiliades2016scaling}. This grid corresponds to the archive, which is also called a ``map''. Each niche only holds the highest-performing individual found so far for this bin, which is called the ``elite''. To create new candidate solutions, parents are selected uniformly among the elites, classic genetic operators are applied, and both the fitness and the features of the offspring are evaluated. The offspring then compete with the current elite of the niche that corresponds to its behavior features: if the niche is empty then the offspring is assigned as the elite of that niche; if the niche is already occupied, then the fitness of the current elite and that of the new candidate are compared, and the best is kept.

A recent study revealed that the elites often occupy a particular part of the genotype space, called the ``elite hypervolume'' \cite{vassiliades:hal-01764739}. Intuitively, this observation means that elites have things in common or that they use similar ``recipes'', so that they are well spread in the feature space but concentrated in the genotypic space. Species in nature follow a similar pattern as they occupy different ecological niches but share a large part of their genomes. For example, fruit flies and humans share 60\% of their genome \cite{fruitfly} while being vastly different animals. When evolving a vector of parameters, this hypervolume can be leveraged by using a variation operator inspired by the cross-over: if two parents are selected from the elite hypervolume, individuals on the line that connects the two parents are likely to be in the hypervolume too, that is, to also be elites (e.g., if the hypervolume is spherical, then any point on the segment that links two random points is also in the sphere). Given two random elites $\mathbf{x}_{i}^{(t)}$ and $\mathbf{x}_{j}^{(t)}$, a new candidate solution $\mathbf{x}_{i}^{(t+1)}$ is generated by:
\label{sec:sota}

\begin{equation}
    \label{eq:line}
  \mathbf{x}_{i}^{(t+1)}=\mathbf{x}_{i}^{(t)}+\sigma_{1} \mathcal{N}(0, \mathbf{I})+\sigma_{2}\left(\mathbf{x}_{j}^{(t)}-\mathbf{x}_{i}^{(t)}\right) \mathcal{N}(0,1)
\end{equation}

\noindent{}where $\sigma_{1}$ and $\sigma_{2}$ are hyperparameters that define the relative strength of the isometric and directional mutations, respectively, and $\mathcal{N}(0,1)$ is the normal distribution with a mean of $0$ and a standard deviation of $1$.

\subsection{Multitask optimization and learning}

To our knowledge, no work in evolutionary computation considers solving thousands of tasks simultaneously. Nevertheless, recent work on ``Evolutionary Multitasking'' \cite{gupta2015multifactorial,da2017evolutionary,hashimoto2018analysis} do attempt to optimize for a few  (typically 2) tasks simultaneously using an evolutionary algorithm. In the most popular algorithm of this family, each individual of the population is assigned a task depending on a ``skill factor'' (an arbitrary number) that is initially chosen randomly, then transmitted to the offspring by randomly assigning the skill factor of one of the parents during cross-over. The tasks are typically weakly related (e.g. Rastringin and Ackley function) and no explicit information from the distance between the tasks is exploited.

Multi-task Bayesian optimization focuses on solving multiple correlated tasks when the fitness function is expensive \cite{pearce2018continuous}, for instance when tuning a machine learning algorithm to several datasets or to tune a policy for a robot that depends on the context \cite{fabisch2014active}, like a walking controller that depends on the slope. The general idea of Bayesian optimization \cite{brochu2010tutorial} is to use the previous fitness evaluations to predict the location of the most promising candidate solution, evaluate it, update the predictor, and repeat. Gaussian processes are usually used to make predictions because they are good at interpolation and they can estimate the uncertainty of the prediction, which is useful in driving exploration.

In multi-task Bayesian optimization, the Gaussian process takes the task coordinates as input, in addition to the candidate solution, which allows the algorithm to predict the performance of every candidate solution on every task, and therefore to choose the most appropriate candidate to evaluate. Overall, multi-task Bayesian optimization solves the same problem as Multi-task MAP-Elites. However, it is tailored to situations in which the fitness function can be called only a few thousand times because the cost to query the Gaussian process is cubic with the number of fitness evaluations. Hence, this algorithm works well only for low-dimensional search spaces (up to six in most papers) and low-dimensional task spaces (up to 5 in \cite{pearce2018continuous}). By contrast, the present work considers tasks that are up to 12 dimensions and candidate solutions that are up to 36 dimensions, but using up to 1 million fitness evaluations.

Lastly, a few algorithms have been proposed in deep reinforcement learning for multi-task learning. In this field, the goal is to learn a single, task-conditionned policy that maximizes the expected return for all the tasks. The general assumption is that a large part of a policy can be re-used accross the tasks (e.g., the visual processing part), and therefore it should be beneficial to learn all the tasks simultaneously. A recent benchmark paper proposed a set of 50 robotic tasks and extended deep reinforcement learning algorithms to multi-task learning (e.g., PPO \cite{schulman2017proximal}, TRPO \cite{schulman2015trust}, SAC \cite{haarnoja2018soft}). The authors report that the best algorithm only solves 36\% of the 50 tasks. Compared to Multi-task MAP-Elites, these algorithms solve fewer but more complex tasks and do not know the correlation between the tasks in advance.

\section{Multi-task MAP-Elites}
\subsection{Problem Formulation}
\label{sec:problem}
Current QD algorithms assume that the fitness function $f(\bm{\theta})$ returns both the fitness value $f_x$ and a feature vector (or behavioral descriptor) $\bm{b}_x$:
\begin{equation}
    f_x, \bm{b}_x \leftarrow f(\bm{\theta})
\end{equation}
By contrast, we are considering a fitness function that is parameterized by a task descriptor $\bm{\tau}$ and returns the fitness value:
\begin{equation}
    f_x \leftarrow f(\bm{\theta}, \bm{\tau})
\end{equation}
The task descriptor might describe, for example, the morphology of a robot or the features of a game level; it is typically a vector of numbers that describes the parameters of the task. In addition, we assume that we have access to a meaningful similarity function between tasks, denoted $d(\bm{\tau_1}, \bm{\tau_2})$.

The overall objective is to find, for each task $\bm{\tau}$, the genome $\bm{\theta}_\tau^*$ with the maximum fitness:
\begin{equation}
    \forall \bm{\tau} \in T, \bm{\theta}_\tau^* = \textrm{argmax}_{\bm{\theta}} \big(f(\bm{\theta}, \bm{\tau})\big)
\end{equation}

Contrary to previous work, we consider domains with many tasks, typically a few thousand.



\subsection{Algorithm}
The Multi-task MAP-Elites algorithm is based on the main principles of MAP-Elites \cite{mouret2015illuminating,vassiliades2016scaling}:
\begin{enumerate}
    \item the diversity space is divided into a large number of niches (or cells) that are organized spatially in an archive (also called ``map'');
    \item each niche contains the best known solution (the elite) for the corresponding combination of features;
    \item to generate a new candidate solution, two random elites are selected from the archive and the traditional cross-over and mutation operators are used.
\end{enumerate}

In Multi-task MAP-Elites, each niche corresponds to a task. The main difference with MAP-Elites is how the niche is determined: in MAP-Elites, this comes for free during the evaluation of the fitness ($\bm{b}_x $, section \ref{sec:problem}), but in Multi-task MAP-Elites, we need to \emph{decide} on which task ($\bm{\tau}$) to evaluate the fitness function.

Intuitively, we want to select a task for which the newly generated candidate has a chance of being fitter than the existing elite.
Our main hypothesis is that two tasks that are close in terms of task distance are likely to also have close solutions; we therefore can choose a task that is close to the niche of one of the parents used to generate the new candidate solution. Choosing a task in the immediate neighborhood of a parent would lead to very limited exploration and was not successful in our preliminary experiments. Instead, we need a \emph{bias} for close tasks while keeping some randomness in the choice to encourage exploration.

Taking inspiration from selection in evolutionary algorithms, we use a \emph{tournament}: we randomly pick $s$ tasks (including tasks for which no elite currently exists), and then from among these, we choose the task closest (in terms of the task distance) to the task of the first parent. This bias is strong when the tournament is large, because we are more likely to pick a task that is close to this parent, and weak when the tournament is small, since we are less likely to pick a close task. In the extreme cases, a tournament of size of the number of tasks corresponds to always taking the task closest to that of the first parent, while a tournament of size 1 corresponds to no bias for proximity at all (i.e., a uniform random choice of the task).

The size of the tournament is therefore critical to the performance of the algorithm. This could be a hyper-parameter, but it is likely to depend on the domain and to require extensive experimentation. Instead, we use a parameter control technique \cite{karafotias2014parameter,eiben1999parameter}, wherein we attempt to identify the best tournament size ``on the fly'' using the data generated since the start of the algorithm. Our measure of success is the number of niches ``invaded'' during a batch of evaluations (e.g., during the last 64 evaluations), that is, how often the newly generated candidate is better than the existing elite for the selected task. For a given tournament, we denote by $r^{(g)}$ the number of successes for a batch of size $B$ (e.g., 64 successive evaluations) at generation $g$ (each generation is a batch), and by $A_{\tau_j}$ the current elite for the task $\tau_j$ that was selected for the $j$-th candidate of the batch:

\begin{equation}
    r^{(g)} = \sum_{j=1, \cdots, B} \mathds{1}_{f(\theta_j, \tau_j) > A_{\tau_j}}
\end{equation}
Where:
\begin{equation}
    \mathds{1}_{f(\theta_j, \tau_j) > A_{\tau_j}} = \left\{
        \begin{array}{ll}
            1 \textrm{ if } f(\theta_j, \tau_j) > A_{\tau_j}\\
            0, \textrm{ otherwise}
        \end{array}
      \right.
\end{equation}

\begin{algorithm}
    \caption{Multi-task MAP-Elites}
    \label{algo:multi}
    \begin{algorithmic}[1]
        \State \Colorbox{\textbf{[Parameters]}}
        \State $T$: vector of tasks
        \State $f(\cdot)$: fitness function
        \State $d(\cdot,\cdot)$: task distance function
        \State \Colorbox{\textbf{[Hyperparameters]}}
        \State $K$: number of random individuals for initialization (e.g., 100)
        \State $E$: total number of evaluations (e.g., $10^5$)
        \State $B$: batch size (e.g., 64)
        \State $S$: (vector) possible tournament sizes (e.g., $[1, 5, 10, 100]$)
        \State \Colorbox{\textbf{[1. Random initialization]}}
        \For{$i \gets 0, K$}
            \State $x \gets \textrm{random\_individual()}$
            \State $\tau \gets \textrm{random\_task(T)}$
            \State $x.fit \gets f(x, \tau)$ \Comment{Evaluate fitness on task $\tau$}
            \State $A[\tau] \gets x$ \Comment{Store $x$ in archive $A$}
        \EndFor{}

        \State \Colorbox{\textbf{[2. Main loop]}}
        \State $s \gets \textrm{random\_in\_list}(S)$
        \Comment{$s$: tournament size}

        \State $\textrm{selected} \gets \textrm{zeros}(\textrm{len}(S))$  \Comment{\# of selections for each size}

        \State $\textrm{successes} \gets \textrm{zeros}(\textrm{len}(S))$
        \Comment{\# of successes for each size}
        \State $e = 0$ \Comment{Evaluation counter}
        \State $g = 0$ \Comment{Generation counter}
        \While{e < E}\Comment{For all the evaluation budget}
            \State $\textrm{selected}[s] \gets  \textrm{selected}[s] + 1$ \Comment{Count selections of $s$}
            \State $g \gets  g + 1$ \Comment{Increase the generation counter}

            \For{$i \gets 0, B$}\Comment{Iterate over a batch}
                \State \Colorbox{\textbf{[2.1 Generate $x$]}}
                \State $p_1 \gets A[\textrm{random\_task(T)}]$\Comment{First parent}
                \State $p_2 \gets A[\textrm{random\_task(T)}]$\Comment{Second parent}
                \State $x \gets \textrm{variation}(p_1, p_2)$\Comment{Mutation \& cross-over}

                \State \Colorbox{\textbf{[2.2 Select the task with a tournament of size $s$]}} \label{algo:tournament}
                \State $\textrm{tasks} \gets s \textrm{ random tasks from } T$  \Comment{Candidates}
                \State $\tau \gets \textrm{closest}(tasks, p_1.task, d)$ \Comment{Tournament}
                \State $x.fit \gets f(x, \tau)$ \Comment{Evaluate fitness on task $\tau$}
                \State $e \gets e + 1$ \Comment{Increase evaluation counter}

                \State \Colorbox{\textbf{[2.3 Try to add $x$ to the archive]}}
                \If{$A[\tau]$ = $\emptyset~\textrm{ or }~x.fit > A[\tau].fit$}
                \State $A[\tau] \gets x$ \Comment{Add $x$ to the archive}
                \State $\textrm{successes}[s] \gets \textrm{successes}[s]  + 1$ \Comment{Count success}
                \EndIf
            \EndFor{}
            \State \Colorbox{\textbf{[2.5 UCB1 algorithm for the tournament size]}} \label{algo:ucb1}
            \State $s \gets S\big[\textrm{arg}\max_{j}\left( \frac{successes[j]}{selected[j]} + \sqrt{\frac{2 \ln(g)}{selected[j]}} \right)\big]$
            \State \Comment{See eq. \ref{eq:ucb}, $j \in 0, \cdots, len(S)$}
        \EndWhile{}
    \end{algorithmic}
    \end{algorithm}

Among the parameter control approaches\cite{karafotias2014parameter}, multi-armed bandits are both straightforward to implement and well founded theoretically. In evolutionary computation, they have, for instance, been successfully used to select the genetic operators \cite{dacosta2008adaptive}. The general idea of multi-armed bandits is to consider each choice (here, each tournament size) as a slot machine with an unknown probability of reward. The objective of the bandit algorithm is to efficiently balance exploitation --- choosing a tournament size that we know can give a good reward --- and exploration --- choosing a size that has not so far been tried very often. It should be noted that the choices in bandit algorithms are \emph{not} ordered, that is, we ignore the fact that a size of 10, for example, is likely to lead to a reward similar to a size of 11. An ordered version of the bandit algorithms would be Bayesian optimization \cite{shahriari2015taking}, but this would be a much more complex algorithm that requires many design choices and much more computation.

One of the simplest and most effective bandit algorithms is UCB1~\cite{auer2002finite}, which achieves the optimal regret up to a multiplicative constant. Given a set of choices organized in a vector $S$ of size $k$ (for instance $S=[1, 10, 100, 1000]$), let us denote by $n_i$ the number of times that $S_i$ has been selected so far. The tournament size $S_g$ at generation $g$ is given by:

\begin{equation}
    \label{eq:ucb}
    S_g = S_{m}, \textrm{where } m = \textrm{arg} \max_{i, \cdots, k}\left(\hat{\mu}_i+ \sqrt{\frac{2 \ln g}{n_i}}\right)
\end{equation}
where $\hat{\mu_i} = \frac{\sum_{j=0,\cdots,g}{r^{(g)}_i}}{n_i}$ is the mean reward (i.e., success) of choice $i$ since the start of the algorithm. When a choice is taken often, $n_i$ increases and $\sqrt{\frac{2 \ln g}{n_i}}$ decreases, which makes it less likely for $S_i$ to be chosen again. At the same time, we tend to choose the tournament size that has a good mean reward (i.e., success) so far (a high value for $\hat{\mu}_i$).

Algorithm~\ref{algo:multi} shows the detailed pseudo-code for Multi-task Map-Elites. It follows the MAP-Elites algorithm with the following exceptions:
\begin{itemize}
    \item a tournament is used for the niche selection (line \ref{algo:tournament});
    \item tournament size is adjusted using UCB1 (line \ref{algo:ucb1}).
\end{itemize}

While Algorithm \ref{algo:multi} is not parallelized, it is easy to do so by generating and evaluating all the candidate solutions of a batch (lines 26--35) in parallel, then attempt to add them to the archive one by one (line 36). This strategy prevents concurrency problems when adding elements and counting successes (for the bandit). Our implementation is in Python and uses this strategy to parallelize the fitness evaluations.

The link to the source code is available in appendix.

\section{Kinematic arm with variable morphology}
\label{sec:arm}
\begin{figure*}
    \begin{center}
    \includegraphics[width=0.94\textwidth]{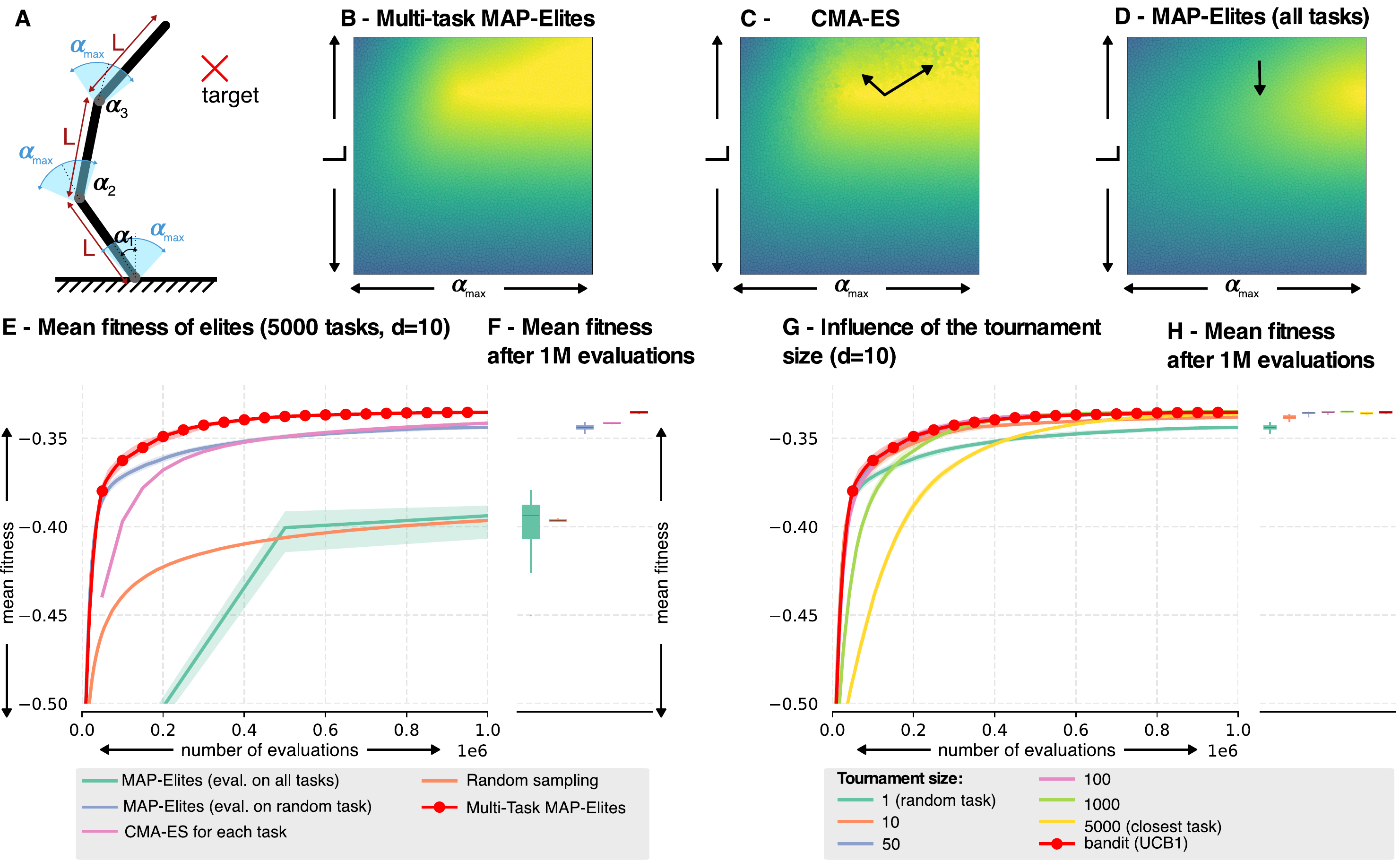}
    \end{center}
    \caption{\label{fig:results_arms}(A) Arm with variable morphology. The objective is to find the angles $\alpha_1,\cdots,\alpha_d$ so that the tip of the arm is as close as possible to the target. The task is parameterized by the link length ($L$), which is the same for each link, and the maximum angular rotation of each joint ($\alpha_{\textrm{max}}$). All the results in this figure are for a 10-link arm. (B) Typical map generated with Multi-task MAP-Elites. (C) Typical map generated when using CMA-ES for each task. (D) Typical map generated with MAP-Elites when evaluating each candidate on each task. (E) Mean fitness over the whole map with respect to the number of evaluations (20 replicates). The solid line represents the median over the replicates and the light zones the interquartile range. (F) Mean fitness over the whole map after 1 million evaluations. All the differences are statistically significant (Mann-Whitney U-test, $p <10^{-10}$), except between MAP-Elites with evaluation on all tasks and random sampling.  (G) Influence of the tournament size on the mean fitness. (H) Mean fitness after 1 million evaluations.}
\end{figure*}

\subsection{Methods}

We first evaluate Multi-task Map-Elites with a planar kinematic arm inspired by previous work \cite{cully2017quality,vassiliades:hal-01764739,cully:hal-01158243} (Fig. \ref{fig:results_arms}-A):
\begin{itemize}
    \item The objective of the task is to find the angle of each joint ($\alpha_1, \cdots, \alpha_d$) so that the tip of the arm (the end-effector) is as close as possible to a predefined target in the plane.
    \item The dimensionality $d$ of the problem is defined as the number of joints (which is equal to the number of links).
    \item By contrast with previous work, we parameterize the arm by the length of the links, denoted $L$ (all the links have the same length), and the maximum angle for each joint, $\alpha_{\textrm{max}}$ (all joints have the same limits).
\end{itemize}
A task $\bm{\tau}$ is defined by a particular combination of $L$ and $\alpha_{\textrm{max}}$ (the task definition is therefore 2-dimensional), and a candidate solution is defined by a vector $\bm{\alpha}=\alpha_1, \cdots, \alpha_d$ ($d$-dimensional), where increasing $d$ increases the task complexity. The joint limits $\alpha_{\textrm{max}}$ and the lengths are normalized by the dimensions $d$ so that the robot has the same total length ($1$ meter) and reaching abilities regardless of the dimensionality:
\begin{eqnarray}
    \alpha_i &=& \frac{1}{d} \cdot (\alpha_i - 0.5) \cdot  \alpha_{\textrm{max}} \cdot  2 \pi\\
    L_\tau &=& \frac{L}{d}
\end{eqnarray}

The fitness function $f(\bm{\alpha}, [L, \alpha_{\textrm{max}}])$ is the Euclidean distance from the tip position $p_d$ to the target position $T$. In these experiments, we arbitrarily set the target to $(1,1)$. The kinematics of the arm can be computed iteratively as follows:
\begin{eqnarray}
M_0 &=& I\\
M_{i+1} &= &M_i \cdot  \left(
    \begin{array}{llll}
        \cos(\alpha_i) & -\sin(\alpha_i) & 0 & L_\tau\\
        \sin(\alpha_i) &  \cos(\alpha_i) & 0 & 0\\
                 0 & 0 &1 & 0\\
                 0 & 0 & 0&1\\
    \end{array}\right)\\
p_{i +1} &=& M_{i+1} \cdot  (0, 0, 0, 1)^T
\end{eqnarray}
Using these notations, the fitness function for a candidate $\alpha$ is the distance between the end of the last link and the target:
\begin{equation}
    f(\bm{\alpha}, [L, \alpha_{\textrm{max}}]) = -\big|\big|p_d - T\big|\big|
\end{equation}
where $\bm{\alpha}$ is the candidate solution, $d$ is the dimensionality of the domain and $T$ the target. Each fitness evaluation requires only $3.5 \times 10^{-4}$ seconds to be evaluated (in Python, Intel(R) Xeon(R) Silver 4110 CPU at 2.10GHz).


\label{sec:arm-baselines}
With an evaluation budget of 1 million, we compare Multi-task MAP-Elites to:
\begin{itemize}
    \item Vanilla MAP-Elites (all tasks): each candidate solution is evaluated on each task and therefore competes in all the niches (each batch therefore requires $B \times card(T)$ calls to the fitness function, that is, the size of the batch times the number of tasks).
    \item Vanilla MAP-Elites (random task): each candidate solution is evaluated on a random task and competes only for that task (in this case, the number of evaluations for a batch is equal to the size of the batch).
    \item CMA-ES \cite{hansen2003reducing}: an independent instance of CMA-ES is run for each task (all the CMA-ES instances run in parallel).
    \item Random sampling: each candidate solution is generated randomly (no cross-over or mutation) and tested on a random task; the best solution is kept as the elite for each task.
\end{itemize}

We define 5000 tasks that are spread evenly using a centroidal Voronoi tesselation \cite{vassiliades2016scaling} (they could also have been spread randomly or on a grid). MAP-Elites uses the line mutation operator introduced in \cite{vassiliades:hal-01764739} (section \ref{sec:sota}, equation \ref{eq:line}). Each batch corresponds to 64 evaluations ($B=64$). We run 30 replicates for each approach to get statistics.

\subsection{Results}

The results (Fig. \ref{fig:results_arms}) show that Multi-Task MAP-Elites outperforms all the baselines. Looking at the generated maps (Fig. \ref{fig:results_arms}-B,D), we observe that CMA-ES leads to some ``noise'' (Fig. \ref{fig:results_arms}-C -- see especially on the top right of the map). This may be caused by the poor performance of some runs (since each run is independent, there is no knowledge transfer at all between tasks) and also, probably, by the fact that with a population of $10$ (the default for CMA-ES in 10 dimensions) and 5000 tasks, only $20$ generations can be run with the evaluation budget (therefore many runs may not have had a chance to converge). The map generated when we evaluate each solution on the 5000 tasks (Fig. \ref{fig:results_arms}-D, bottom) is smoother, but the quality of the solutions is much lower than in Multi-task MAP-Elites. This can be explained by the fact that only a few generations can be completed with 1 million evaluations: only 3 batches (192 solutions, since $64\times5000=320,000$) have been evaluated within the budget.

Contrary to classic MAP-Elites experiments, all the niches in the experiments conducted here were reachable by any candidate solution. As a result, all the approaches tested here were able to quickly fill all the niches, making map-coverage a moot point for comparison. We therefore focus our analysis on the mean fitness over the map (Fig. \ref{fig:results_arms}-E,F). Both random sampling and evaluation on all the tasks lead to low fitness values. MAP-Elites on random tasks performs as well as CMA-ES, which might be surprising at first since CMA-ES is one of the best known algorithms for continuous optimization. However, this good performance can be explained by the fact that even when MAP-Elites selects the task randomly, it creates new candidate solutions from the elites of the map: if the best solutions of all the niches share some genes (they use ``the same recipe''), then they occupy the same ``elite hypervolume'' \cite{vassiliades:hal-01764739} and the line mutation operator can leverage a good solution from one task to solve another one. Nevertheless, Multi-task MAP-Elites quickly outperforms (in a few hundreds of evaluations) both CMA-ES and MAP-Elites.

In order to investigate the influence of the bandit algorithm, we also ran Multi-task MAP-Elites with a fixed tournament size, using values ranging from 1 (equivalent to selecting a random task) to 5000 (equivalent to selecting the closest task). The results (Fig.\ref{fig:results_arms}-G,H) show that, a tournament that is too small (less than 50) does not allow the algorithm to reach the best mean fitness values. Conversely, a large tournament (more than 1000) slows down the algorithm significantly, but it does not prevent it from reaching the best fitness values. The bandit algorithm finds the best tradeoff as it is as good as the best tournament value.

\section{Six-legged Locomotion}

\begin{figure*}
    \centering
    \includegraphics[width=0.94\textwidth]{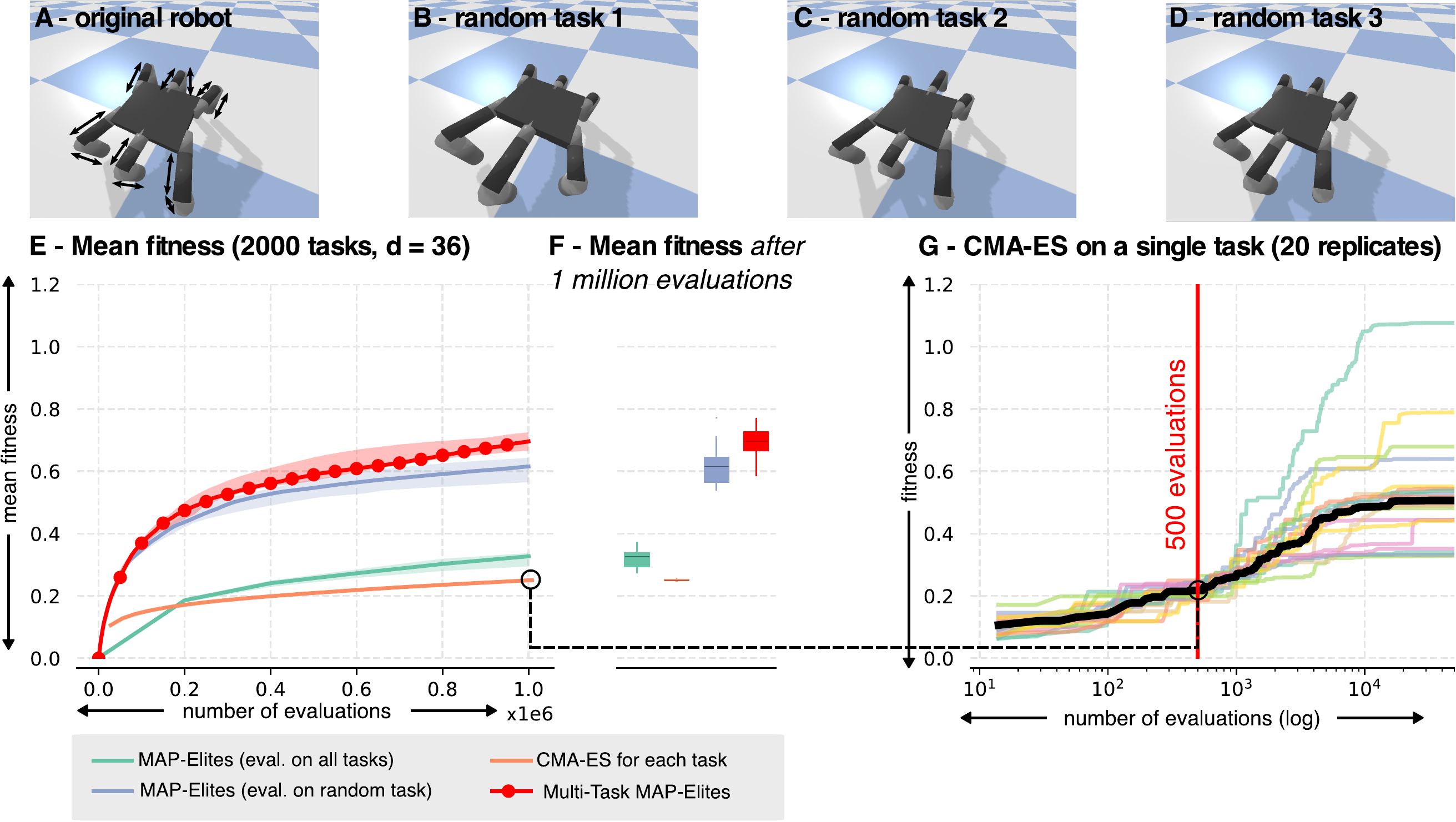}
    \caption{\label{fig:results_hexapod}Hexapod robot experiments. (A) The robot is parameterized by 12 values that correspond to the length of the 12 segments indicated by arrows. (B-D) Examples of randomly generated morphologies. (E) Mean fitness for all tasks with respect to the number of evaluations (20 replicates). The solid line represents the median over 20 replicates and the light zones the interquartile range. (F) Mean fitness value after 1 million evaluations. All the differences are statistically significant (Mann-Whitney U-test, $p < 0.0005$). (G) Fitness of gaits generated by CMA-ES if given many more evaluations (here, 100,000) and a single task. Each line is an individual run and the thick black line is the median.}
\end{figure*}
\subsection{Methods}
To evaluate Multi-task Map-Elites in a more challenging scenario, we use a simulated 6-legged robot \cite{cully:hal-01158243,chatzilygeroudis:hal-01654641} that is required to walk forward as fast as possible. The morphology of the 6-legged robot is parameterized by 12 values that correspond to a length change for each of the 12 main segments (2 segments for each leg, see Fig. \ref{fig:results_hexapod}-A). Each set of lengths defines a task since a specific gait is likely to be needed for a specific morphology, however, a given gait might work or be a good starting point for several morphologies.

The main differences from the previous experiments are that: (1) the problem is harder (experiments in this domain show that finding a good gait requires thousands of evaluations \cite{cully:hal-01158243}), (2) there are more parameters to optimize ($36$ versus $10$), and (3) the tasks are defined by more parameters ($12$ versus $2$), (4) the tasks are defined randomly instead of being spread with a centroidal Voronoi tesselation. Simulating a walking 6-legged robot for 3 seconds is about 3 orders of magnitude more computationally demanding than computing the forward kinematics of a planar arm.

Finding optimal gaits for many variants of the same morphology could be useful in future work for damage recovery. In particular, it was previously shown that a repertoire of diverse gaits generated for the same
morphology can be combined with Bayesian optimization to allow a 6-legged robot to recover from damage in a few minutes \cite{cully:hal-01158243,pautrat2017bayesian}. Put differently, damage conditions were not explicitly anticipated but the diversity of gaits was enough to find compensatory gaits. By contrast, the repertoire generated with Multi-task MAP-Elites explicitly anticipates damage conditions (here, different lengths for the leg segments) --- which could help for recovery. Nevertheless, the same Bayesian optimization as in previous work \cite{cully:hal-01158243,pautrat2017bayesian} can be applied to this problem as well.

Incidentally, searching for optimal gaits for a large set of morphlogies is an original approach to co-evolve gaits and morphology, which is a classic line of research in evolutionary robotics \cite{doncieux:hal-01131267}. Thus, by looking at the highest-performing gaits in a map, we find the highest-performing morphology/gait pair.

The gaits being used here are generated by the same 36-dimen\-sional controller used in previous work \cite{cully:hal-01158243,pautrat2017bayesian,chatzilygeroudis:hal-01654641}. In a few words, the robot has 18 joints (3 per leg). 12 joints are explicitly controlled (2 per leg) and the last joint in each leg is equals to that for the second joint (so that the last segment stays mostly vertical). The trajectory of each joint is a periodic signal (a smoothed square wave) that is parameterized by a phase shift, an amplitude, and an offset. There are therefore 3 parameters for each of the 12 controlled joints.

The robot is simulated in Pybullet\footnote{Source code: \url{https://github.com/resibots/pyhexapod}}~\cite{coumans2016pybullet}. In these simulations, the robot is initially located at the origin, $(0, 0)$, and is supposed to move along the $x$-axis.

The fitness function $f(\bm{\theta}, \bm{\tau})$ is the distance covered along the $x$-axis until one of the following conditions is met:
\begin{itemize}
    \item the simulation lasts more than 3 simulated seconds;
    \item the absolute value of the pitch or roll of the body exceeds $\frac{\pi}{8}$ (indicating that the body is not horizontal enough);
    \item the absolute value of the $y$ position is above $0.5$ (indicating that the robot has deviated too much from a straight path).
\end{itemize}
Each fitness evaluation requires about $0.3$ seconds to be evaluated (in Python, Intel(R) Xeon(R) Silver 4110 CPU at 2.10GHz).

In these experiments, we generate 2000 random morphologies, that is, 2000 tasks. The distance between tasks is the Euclidean distance between their 12 parameters. As in the previous experiments with the kinematic arm, we use 1,000,000 evaluations and a batch size of $64$. Besides the number of tasks and the dimensionality, all the parameters are the same as in the previous experiments. We replicate each experiment 20 times to gather statistics.

We use the same baselines as before (section \ref{sec:arm-baselines}): running 2000 instances of CMA-ES in parallel (one for each task), using MAP-Elites with random task assignment, and using MAP-Elites with an evaluation of each offspring on all the tasks.

\subsection{Results}
As with the kinematic arm experiments, the results show that Multi-task MAP-Elites outperforms the baselines (Fig. \ref{fig:results_hexapod}). MAP-Elites with random task assignment is competitive with Multi-task MAP-Elites, which confirms that MAP-Elites can leverage the elite hypervolume for multi-task optimization. Surprisingly, CMA-ES is outperformed by all the other approaches --- even MAP-Elites with evaluation on all the tasks. The most likely causes for this low performance are (1) with 2000 tasks and a default population of 14 (the population size computed by CMA-ES for 36 dimensions), only 500 evaluations are used for each instance, whereas benchmarks for black-box optimization usually use more than $10^6$ evaluations for a single optimization~\cite{hansen2010comparing}); and (2) the 6-legged locomotion task is much harder for CMA-ES than the kinematic arm because the gradient is less smooth and requires a more global optimization.

\begin{figure*}
    \centering
    \includegraphics[width=0.95\textwidth]{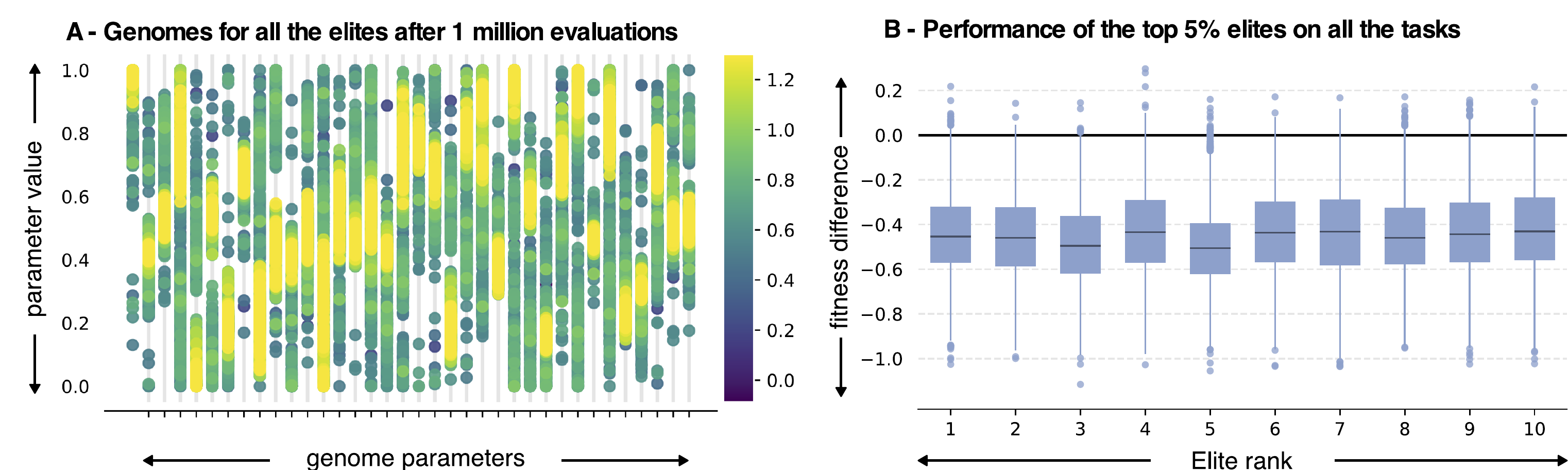}
    \caption{\label{fig:analysis_hexapod}(A) Genomes of a typical map after 1 million evaluations. Each column is a parameter of the genome; the vertical coordinate corresponds to the parameter value, and the color to the fitness value. (B) Fitness of the top 5\% of elites on all the tasks compared to the fitness of the elite for the task (< 0 indicates that the elite in the map performs better than the top-5\% elite; > 0 indicates that the top-5\% elite performs better than the current elite in the map).}
\end{figure*}

To better understand the performance of CMA-ES in our case, we ran 20 replicates of vanilla CMA-ES for 100,000 evaluations on a single task (randomly chosen). The results show that CMA-ES can find high-performing solutions, but there is a large variance between the runs, as only 4 replicates out of 20 found a fitness above 0.7 (the median mean value found by Multi-task MAP-Elites in 1 million evaluations). In addition, most replicates stop improving after about $10,000$ evaluations, which indicates a premature convergence. Overall, these results show that Multi-task MAP-Elites benefits from solving all the tasks simultaneously, both to avoid premature convergence and to find high-performing solutions faster. It should be noted that more recent versions of CMA-ES (e.g., Bipop CMA-ES) could perform better, but it is unlikely that they can find high-performing gaits in only 500 evaluations.

We analyzed the genomes of a typical map generated by Multi-task MAP-Elites with 1 million evaluations (Fig. \ref{fig:analysis_hexapod}-A). The plots show that the values for each parameter tend to be in a specific range, often around the parameters that correspond to the best fitness. For instance, in this map, there is no elite with a value around 0 for the first parameter, and most of the elites have a value between 0.6 and 1.0. This shows that the elites follow common patterns (at least in terms of ranges for each parameter), which explains why MAP-Elites and Multi-task MAP-Elites are effective: they can leverage these patterns to generate new individuals.

Lastly, we also checked that the tasks were different enough so that the best gait needs to be different for each task. To do so, we evaluated the fitness of the top 5\% of elites (i.e., top-10 elites) on  all the tasks and we compared the obtained fitness with that for the elite found by Multi-task MAP-Elites for each task. A negative value means that the elite found by Multi-task MAP-Elites for this task is better than the top-5\% elite, indicating that a simple copy of the elite to all the tasks would be ineffective. On the other hand, a positive value means that the top elite performs better than the elite found for that task. The results (Fig. \ref{fig:analysis_hexapod}-B) show that the fitness difference for these elites all lie at about -0.4, indicating that the best elites do not perform well on the other tasks. A corollary is that the best fitness values of the map are likely to correspond to better morphology (e.g., symmetric morphologies), and not to a better optimization of the gait. Our preliminary analysis tends to show that the best morphology corresponds to large middle legs, and short front and back legs.

\section{Conclusion}
This paper shows that there is a strong connection between quality diversity algorithms and multi-task optimization, especially when a few thousands tasks are considered. MAP-Elites with random task assignment is a straightforward algorithm and performs surprisingly well in the two domains that we considered: the performance is similar to CMA-ES (a conceptually much more complex algorithm) in the 10-dimensional arm / 5000 case and much better than CMA-ES in the hexapod case. Our interpretation is that MAP-Elites with the line mutation operator effectively exploits the elite hypervolume to fill niches with high-performing solutions. Put differently, when the tasks are assigned randomly, MAP-Elites exploits the similarity between the tasks implicitly.

Nevertheless, the bias introduced by Multi-task MAP-Elites improves the results substantially and does not add any hyperparameter, thanks to the bandit algorithm. A natural follow-up will be to investigate whether maps generated by varying the morphology of robots (i.e., what is done here) are better for damage recovery than maps generated by encouraging different behaviors with the same morphology (i.e., what has been done so far, for example in \cite{cully:hal-01158243}).

The main drawbacks of Multi-task MAP-Elites are that it assumes that (1) there is at least few hundred tasks (with fewer tasks, there is not enough diversity for MAP-Elites to work), and (2) we have access to a distance function between tasks. In complex cases, for instance, game levels, this distance might not be available\footnote{Some work has been published recently for defining or learning distance functions between game levels \cite{smith2010analyzing,summerville2018expanding,justesen2018illuminating}, although we did not test them in combination with Multi-Task MAP-Elites so far.}, which prevents a direct use of Multi-task MAP-Elites. In these cases, using MAP-Elites with random task assignment is a viable option. Still, future work should investigate how to discover or learn this distance function during evolution~\cite{smith2010analyzing,summerville2018expanding,justesen2018illuminating}.

Overall, future work should further explore the connections between QD algorithms and multi-task optimization, both to import novel ideas to QD algorithms and to propose new algorithms for the multi-task optimization and learning communities.

\section*{Source code}
Reference implementation of Multi-task MAP-Elites (Python3): \url{https://github.com/resibots/pymap_elites}.\\
Experiments: \url{https://github.com/resibots/2020_mouret_gecco}

\section*{Acknowledgements}
This work received funding from the European Research Council (ERC) under the European Union's Horizon 2020 research and innovation programme (GA no. 637972,
project ``ResiBots'') and the Lifelong Learning Machines program (L2M) from DARPA/MTO under Contract No. FA8750-18-C-0103.

\bibliographystyle{ACM-Reference-Format}
\bibliography{multitask.bib}

\end{document}